%% file: main.tex
\documentclass[sigconf, authorversion]{acmart}  %

\usepackage{comment}
\usepackage{color}
\usepackage{setspace}
\usepackage[noend]{algorithmic}
\usepackage{algorithm}
\usepackage{graphicx}
\usepackage{geometry}
\usepackage{subfig}
\usepackage{textcomp}
\usepackage{caption}
\usepackage{enumitem}
\usepackage{url}
\usepackage{xcolor,colortbl}
\usepackage{balance}
\usepackage{cleveref}

\AtBeginDocument{%
  \providecommand\BibTeX{{%
    \normalfont B\kern-0.5em{\scshape i\kern-0.25em b}\kern-0.8em\TeX}}}

\setcopyright{cc}
\acmConference[Under Review]{Under Review}{}{}
\acmBooktitle{Under Review}

\begin{document}

\title{A Privacy-Preserving Hybrid Federated Learning Framework for Financial Crime Detection}

\author{Haobo Zhang}
\affiliation{%
  \institution{Michigan State University}
  \streetaddress{426 Auditorium Road}
  \city{East Lansing}
  \state{Michigan}
  \country{USA}}
\email{zhan2060@msu.edu}

\author{Junyuan Hong}
\affiliation{%
  \institution{Michigan State University}
  \streetaddress{426 Auditorium Road}
  \city{East Lansing}
  \state{Michigan}
  \country{USA}}
\email{hongju12@msu.edu}

\author{Fan Dong}
\affiliation{%
  \institution{University of Calgary}
  \streetaddress{2500 University Drive NW}
  \city{Calgary}
  \state{Alberta}
  \country{Canada}}
\email{fan.dong@ucalgary.ca}

\author{Steve Drew}
\affiliation{%
  \institution{University of Calgary}
  \streetaddress{2500 University Drive NW}
  \city{Calgary}
  \state{Alberta}
  \country{Canada}}
\email{steve.drew@ucalgary.ca}

\author{Liangjie Xue}
\affiliation{%
  \institution{Coinbase Global, Inc.}
  \streetaddress{}
  \city{}
  \state{}
  \country{USA}}
\email{liangjie.xue@gmail.com}

\author{Jiayu Zhou}
\affiliation{%
  \institution{Michigan State University}
  \streetaddress{426 Auditorium Road}
  \city{East Lansing}
  \state{Michigan}
  \country{USA}}
\email{jiayuz@msu.edu}

\renewcommand{\shortauthors}{Zhang, et al.}
\newcommand{\comm}[1]{\textcolor{red}{#1}}

\input{sec/0_Abstract.tex}



\maketitle

\input{sec/1_Intro.tex}

\input{sec/2_Background.tex}

\input{sec/3_Overview.tex}

\input{sec/4_ThreatModel.tex}

\input{sec/5_System.tex}
\input{sec/6_ExperimentalResults.tex}

\input{sec/9_Conclusion.tex}

\vspace{-0.1in}
\subsection*{Acknowledgment}
This material is based in part upon work supported by the
National Science Foundation under Grant IIS-2212174, IIS-1749940, Office of Naval Research
N00014-20-1-2382, and National Institute on Aging (NIA) RF1AG072449.
This research was supported in part by the University of Calgary Start-up Funding 10032260.

\newpage
\bibliographystyle{ACM-Reference-Format}
\bibliography{reference}

\input{sec/10_Appendix.tex}
\end{document}

%% file: sec/0_Abstract.tex
\begin{abstract}
The recent decade witnessed a surge of increase in financial crimes across the public and private sectors, with an average cost of scams of \$102m to financial institutions in 2022. 
Developing a mechanism for battling financial crimes is an impending task that requires in-depth collaboration from multiple institutions, and yet such collaboration imposed significant technical challenges due to the privacy and security requirements of distributed financial data. 
For example, consider the modern payment network systems, which can generate millions of transactions per day across a large number of global institutions. Training a detection model of fraudulent transactions requires not only secured transactions but also the private account activities of those involved in each transaction from corresponding bank systems. 
The distributed nature of both samples and features prevents  most existing learning systems from being directly adopted to handle the data mining task. 
In this paper, we collectively address these challenges by proposing a hybrid federated learning system that offers secure and privacy-aware learning and inference for financial crime detection.  
We conduct extensive empirical studies to evaluate the proposed framework's detection performance and privacy-protection capability, evaluating its robustness against common malicious attacks of collaborative learning. 
We release our source code at \url{https://github.com/illidanlab/HyFL}.

\end{abstract}

%% file: sec/1_Intro.tex
\section{Introduction}

The recent decade has witnessed increasing occurrences and types of financial crime. According to a U.S. report in 2022, 62\% of all surveyed financial institutions (FIs) experienced an increase in financial crime, and the average cost of scams to each FI surveyed is \$102m from 2021 to 2022~\cite{featurespace}. The overwhelming financial crimes have greatly harmed industries and made them hesitant to innovate. While blockchains have been demonstrated as a revolutionary tool for finance infrastructure and innovation, financial crimes are commonplace, with \$3.5b sent from criminal Bitcoin addresses in 2020 alone and \$41.2m sent to criminals directly from US exchanges~\cite{ciphertrace}.

To battle against financial crimes, many infrastructures and tools have been built over the years. For example, the U.S. Department of the Treasury has set up Financial Crime Enforcement Network (FinCen) to protect financial systems from money laundry and collect information for the defense against financial crime~\cite{fincen}. The Federal Reserve published an intuitive classifier called FraudClassifier~\cite{fedpaymentsimprovement}. Despite the efforts to safeguard, criminals have kept improving the means of financial crimes by more subtle approaches that are harder to detect by conventional methods. 

\begin{figure}[t]
    \centering
	\includegraphics[width=0.49\textwidth]{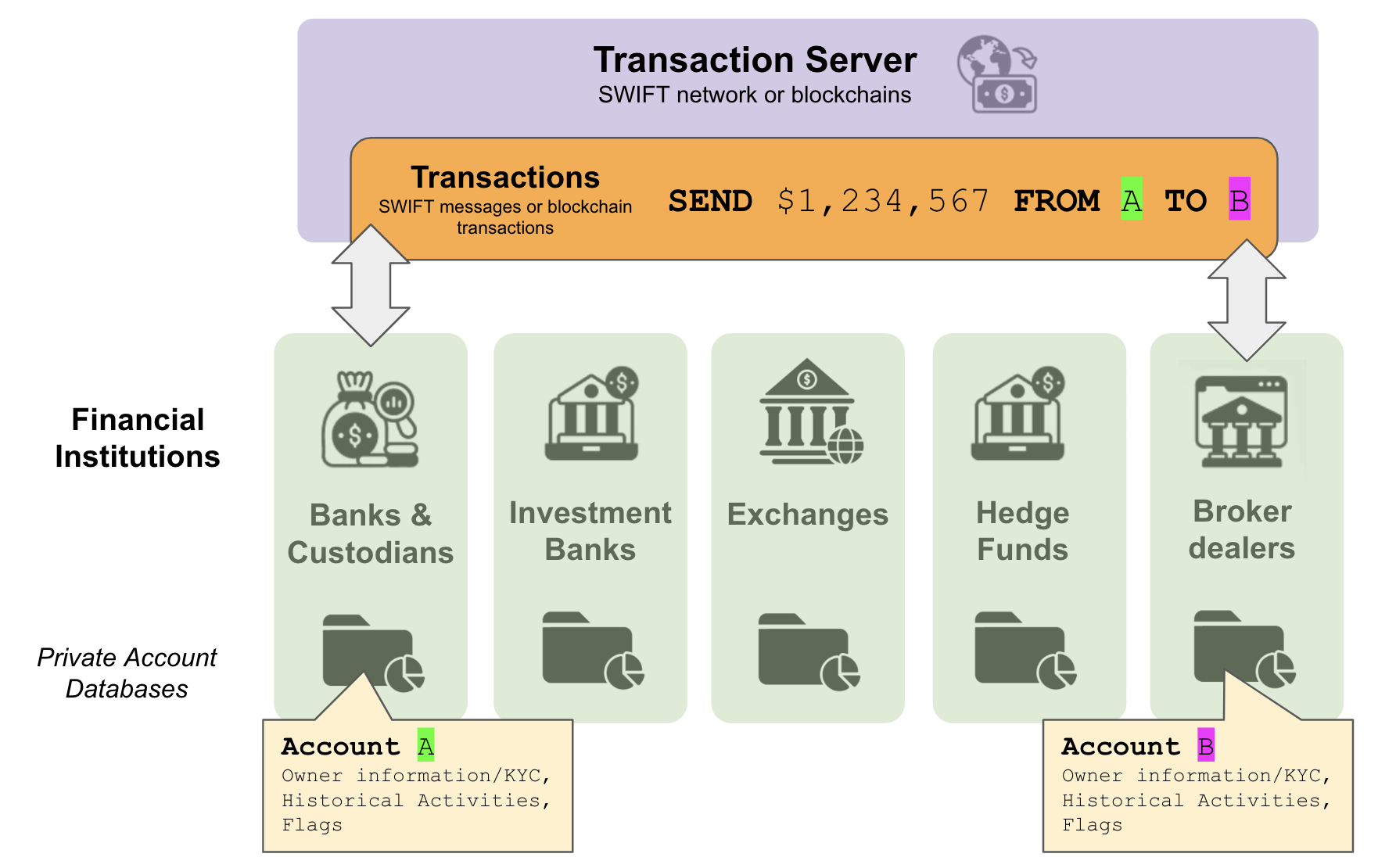}
	\caption{Financial crime detection needs collaborative efforts from multiple financial institutions. A typical transaction moves funds between two accounts in two financial institutions. To determine if a transaction is fraudulent or not, we rely on a collective analysis of the information associated with the transaction and information of the source and target accounts at different institutions. Building models require learning and inference across private and distributed databases in different institutions, which imposes significant challenges and demands a novel learning paradigm. }
	\label{fig:HyFL}
 \vspace{-10pt}
\end{figure}

More recently, financial institutions have sought to use artificial intelligence and machine learning to analyze real-time data to identify financial crimes and have demonstrated this to be very promising~\cite{rouhollahi2021towards}, where 86.3\% of financial institutions have placed machine learning or AI as a high priority on innovation or improvements. Indeed, modern AI, backed up by machine learning and data mining, has demonstrated the exceptional capability of detecting complicated patterns from data. Yet to fully leverage the powerful modeling capability, the AI system needs to access a variety of data, including samples and features, to pick up patterns. 
For example, consider the modern payment network systems, which can generate millions of transactions per day across a large number of global institutions. Training a detection model of fraudulent transactions requires not only secured transactions but also the private account activities of those involved in each transaction from corresponding bank systems. 
Another example is detection modeling for blockchain transfers. The analysis of blockchain transactions can benefit from account activities associated with the crypto addresses involved in the transaction. 

As such, data-driven detection models of financial crimes require the collaboration of multiple financial institutions during both the training and inference stages. 
Again, consider the case of detection in payment network systems: wire transactions are stored in a server, and account activities are stored in individual bank databases. In order to analyze one wire transaction sending funds between two banks, we need jointly analyze the transaction information from the server, account activities of the source account from the bank of the source account, and also activities of the target account from another bank. 

Both transactions and bank account activities are very sensitive data owned by different financial institutions, and learning a model requires privacy-aware collaborative learning. With the increasing demand and law regulation (e.g., General Data Protection Regulation) for learning from distributed data without directly sharing them, the federated learning paradigm has been developed and attracted great efforts from the data mining and machine learning community. The most common type is horizontal federated learning, in each iteration of which a central server retrieves updated models from participating clients using partial samples and then sends the aggregated model to them for iterative updates. Vertical federated learning, on the other hand, handles the setting where clients have different parts or features of the same set of samples. However, the distributed and private nature of \emph{both samples and features} prevents most existing learning systems from being directly adopted to handle the financial crime prediction task. Besides the learning stage, the inference stage of the model also needs private information from distributed parties. 

In this paper, we develop a novel and holistic privacy-preserving approach for financial crime detection from distributed and private financial information. To achieve this goal, the proposed solution leverages a hybrid of vertical and horizontal federated learning: From the vertical perspective, the \emph{transaction client} (e.g., the server) will share extracted features with \emph{account clients} (e.g., banks) for collaborative prediction; from the horizontal perspective, the bank units will collaboratively train a unified encoder to extract features from private account information. 
We combine them by fusing the features from the two involved parties and leveraging new features to train a predictive model in the transaction client. 
The proposed solution considers comprehensive privacy risks and associated attacks, including model inversion~\cite{fredrikson2015model}, privacy attribute inference~\cite{jia2018attriguard}, membership inference~\cite{shokri2017membership}, and feature leakage. To protect against these risks, the proposed system combines strategies including noise injection, local feature extraction, and encryption technique. 

Extensive empirical results on the anomaly detection task show that the proposed hybrid federated learning approach achieves a great balance between model performance and privacy protection. As such, it provides a powerful financial crime detection tool to securely and privately conduct collaborative data analysis from multiple financial institutions. Also, the proposed collaborative framework provides a powerful tool for law enforcement to conduct financial crime detection, while protecting the privacy of participating financial institutions.

%% file: sec/2_Background.tex
\vspace{-0.1in}
\section{Background}

\noindent\textbf{Financial crime detection.} 
Detecting financial crime has been an ongoing challenge, where the goal is to identify abnormal transactions, including fraud transactions, anomalous payments, and money laundering~\cite{kurshan2020graph, kurshan2020financial, rouhollahi2021towards}. As money transfer transactions involve multiple parties, their sophisticated patterns of evolving financial crimes demand joint data analysis from multiple financial institutions.
For instance, an abnormal transaction may be sent to a frozen account. Analyzing this abnormal transaction requires the account status (frozen or active) which is stored in a bank, as well as the transaction record from the transaction client.
To develop a data-driven financial crime detection system, a machine learning model has to combine the features from the two systems in order to predict crime behaviors precisely.
However, data records in the transaction client and banks are located distributedly and cannot be shared due to their sensitive nature. As such, it is challenging to combine the information sources.
To learn an effective model, transaction clients and bank servers need to collaborate to extract knowledge from millions of transactions and the associated accounts.

\noindent\textbf{Distributed federated learning.} 
Due to growing concerns about privacy leakage from sharing personal data, privacy regulations (e.g., GDPR~\cite{gdpr} in Europe) has been enforced in recent years, prohibiting user data from being transmitted to a third party.
To enable distributed machine learning that conforms to data privacy regulations, federated learning (FL) does not require data to be directly shared but instead trains a model collaboratively from distributed clients (data sources). 
The main idea of federated learning is to aggregate knowledge without sharing raw data from multiple clients through locally-trained models~\cite{yang2019federated}.
FL has gained traction in multiple domains, including computer vision~\cite{he2021fedcv, liu2020fedvision}, graph learning~\cite{lalitha2019peer, he2021fedgraphnn}, natural language understanding~\cite{lin2021fednlp}, and etc.
Depending on the status of feature and data splitting, FL can be further categorized into two classes: vertical and horizontal federated learning~\cite{ZHANG2021106775, yang2019federatedconcept}.
Vertical federated learning learns a model that predicts targets based on features concatenated from different clients with synchronized sample indexes.
On the other hand, horizontal federated learning maintains the same feature space while collecting non-overlapped data from clients.

\noindent\textbf{Challenges of existing federated learning.}
Existing FL frameworks cannot be directly adopted in the collaborative learning task of financial crime detection due to their hybrid nature: the vertical relation between the \emph{account clients} and \emph{transaction clients} and the co-existing horizontal relation between different banks.
Account clients and transaction clients need to share features for prediction, but the corresponding account client is joined based on the account identification (e.g., an account number or a public key address in blockchains) in transactions.
Meanwhile, account clients have to share knowledge such that a more powerful model can be trained.

\noindent\textbf{Privacy risks in federated crime detection.}
Like other FL frameworks, there are non-neglectable privacy risks in hybrid FL on data sharing and model publishing.
We consider the following four types of privacy risks in this paper: 
\emph{Model inversion:} model inversion leaks information reversing the final trained model~\cite{fredrikson2015model}. By setting up an appropriate loss, optimization-based methods can be used to estimate the original input.
\emph{Attribute inference:} the correlations between features can be leaked via attribute inference with the support of some accessible features like some public data~\cite{gong2016you, jia2018attriguard}.
\emph{Membership inference:} while one can infer whether or not a given data point is present in the training set using membership inference~\cite{hu2021membership, shokri2017membership}.
\emph{Feature leakage:} extracted features can also be leaked from untrusted units to third parties, i.e., the feature leakage risk. 

\noindent\textbf{Mitigation of privacy risks.}
There have been three main-stream strategies to defend against privacy risks. 
\emph{Differential privacy (DP)} adds Gaussian noise or Laplacian noise to the gradients during the training process~\cite{abadi2016deep, truex2020ldp} to defend various attacks. Then the differential privacy property guarantees that it is difficult to distinguish two adjacent datasets.
\emph{Homomorphic encryption (HE)} ensures that users can perform computation on encrypted data without first decrypting it~\cite{acar2018survey, fontaine2007survey}, with which two nodes can operate the encrypted data without leaking the original data to the third party.
\emph{Adversarial privacy-disentanglement (ADV)} aims to adversarially remove the known privacy attribute from the learned representations while preserving the utility attributes~\cite{wu2018towards, wu2020privacy}.
However, DP and ADV lead to the degradation of model performance, while HE suffers from high computation costs and limited computation operation.

\vspace{-10pt}
\section{Proposed Method}

To address the challenges in the previous section, we hereby propose a novel \emph{Hybrid Federated Learning} (HyFL) framework that integrates the vertical and horizontal frameworks smoothly.

\begin{figure}[t]
    \centering
	\includegraphics[width=0.45\textwidth]{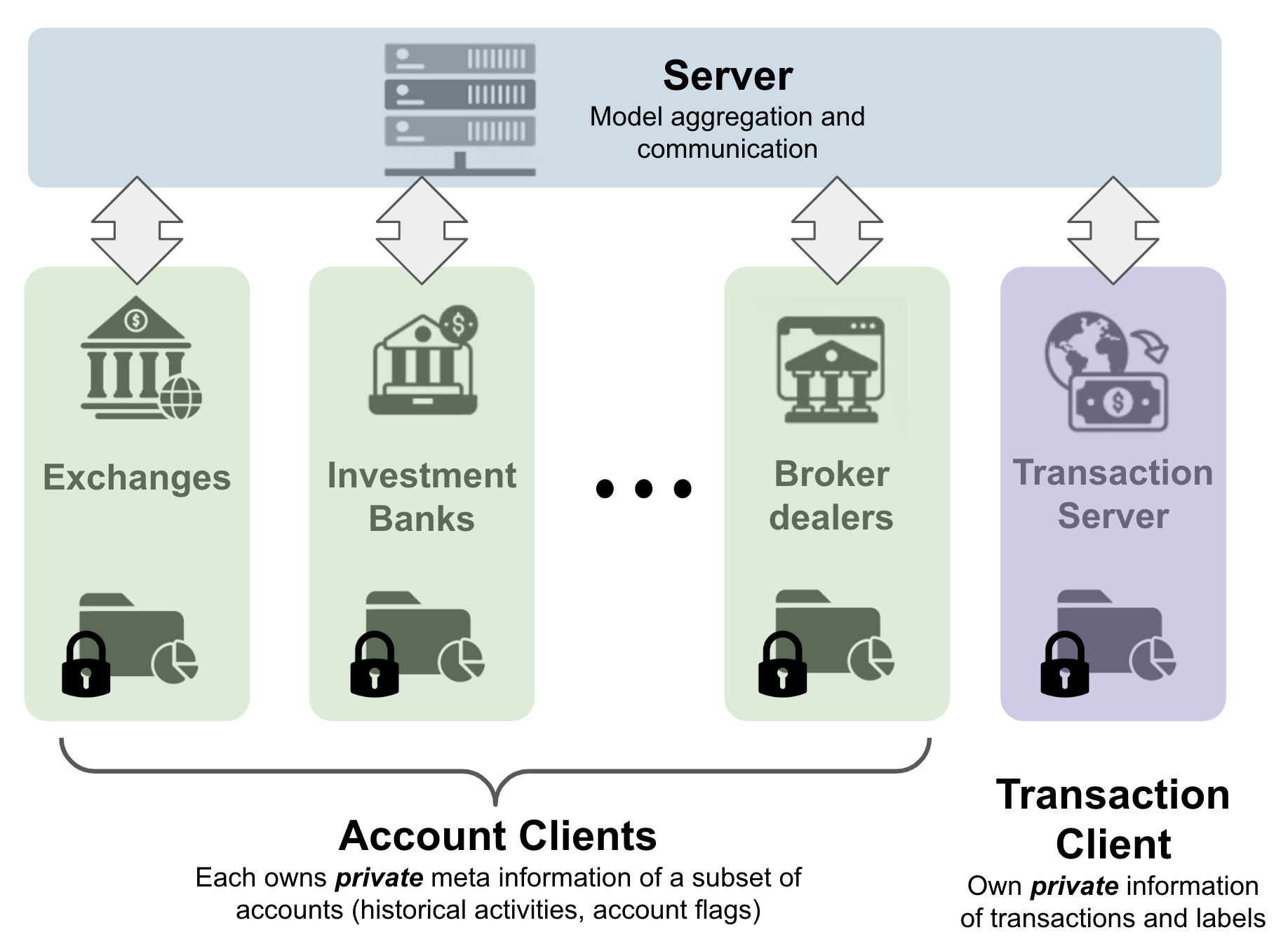}
 \vspace{-0.1in}
	\caption{The federated setup and three key participants of the proposed HyFL framework: 1) A \emph{server} responsible for information aggregation; 2) \emph{account clients}: a set of financial institutions owning meta information of accounts; 3) A \emph{transaction client} that owns private transactions and their labels (abnormal or not).}
	\label{fig:HyFL:setup}
  \vspace{-0.10in}
\end{figure}

For simplicity of discussion without the loss of generality, we assume that one \textbf{transaction client} (Tx client) has all \emph{private} transaction data, and the \emph{private} account metainformation is stored in a set of \textbf{account clients} (Ac clients) separately.
During the training phase, the label information (whether or not a transaction is marked as fraud) is stored in the transaction client with an option to be stored in the aggregation server as well. 
In that case, since the data in account clients share the same dimensions of features, they can be viewed as the clients in a horizontal FL setting.
Compared to the transaction client, account clients have different features even for the same transaction. 
As a result, a vertical FL setting will apply between the account clients and transaction clients.
We illustrate the federated setup and key components in Figure~\ref{fig:HyFL:setup}.

\noindent\textbf{Notations.}
We now introduce the notations to be used in this paper. 
Formally, consider a set of $M+1$ clients $\{1,2,...,M+1\}$ with $M$ account clients and one transaction client.
Define a dataset $\mathcal{D}=\{(\mathbf{x}_i,y_i)\}_{i=1}^N$.
In the vertical FL setting, each client $m$ has a subset of the features from $\mathcal{D}$, i.e., $\mathcal{D}_m = \{ \mathbf{x}_{i,m} \}_{i=1}^N$, where $\mathbf{x}_{i,m}$ contains the features of $\mathbf{x}_i$ stored in the client $m$. 
The label set $\{y_n\}_{i=1}^N$ can be stored on either a client or the server.
In the horizontal FL setting, each client $m$ has a partition of the dataset, i.e., $\mathcal{D}_m = \{ (\mathbf{x}_i, y_i) \}_{i=1}^{N_m}$, where $N_m$ is the size of data stored in client $m$. The label set is stored with the data on each client in horizontal settings.

%% file: sec/3_Overview.tex
\subsection{Framework Overview}
\label{sec:overview}

In this section, we give an overview of our proposed framework. 
First, we define three types of computation nodes, based on which we propose our hybrid federated learning framework. 
Then we introduce three phases in the training stage, where an auto-encoder is trained on the account data and used to extract the feature embedding. Then a classifier is trained in the transaction client to detect potential financial crime.

\noindent\textbf{Computation Nodes and Learning Framework.}
In the proposed system, we coordinate two types of computation nodes and a server for hybrid federated learning.
\begin{enumerate}[leftmargin=*]
    \item The \textbf{transaction client} consists of a set of features describing the transactions for the identification of anomalies in our data, as well as the labels indicating whether a transaction involves a crime or not. 
    With the information, the transaction client is able to train a classifier locally and independently and store the classifier parameters locally.
    \item The \textbf{account clients} maintain account activities with their corresponding meta-information (e.g., account flags).  The account information provides complementary information for detecting anomalies, as many criminal behaviors are related to historical account activities. We use an autoencoder as a feature extractor of flags in account clients.
    \item A computation \textbf{server} aggregates the feature extractors from the account clients and sends the feature embedding from the account clients to the transaction client.
\end{enumerate}

\begin{figure*}[t]
\centering
    \includegraphics*[width=0.75\textwidth]{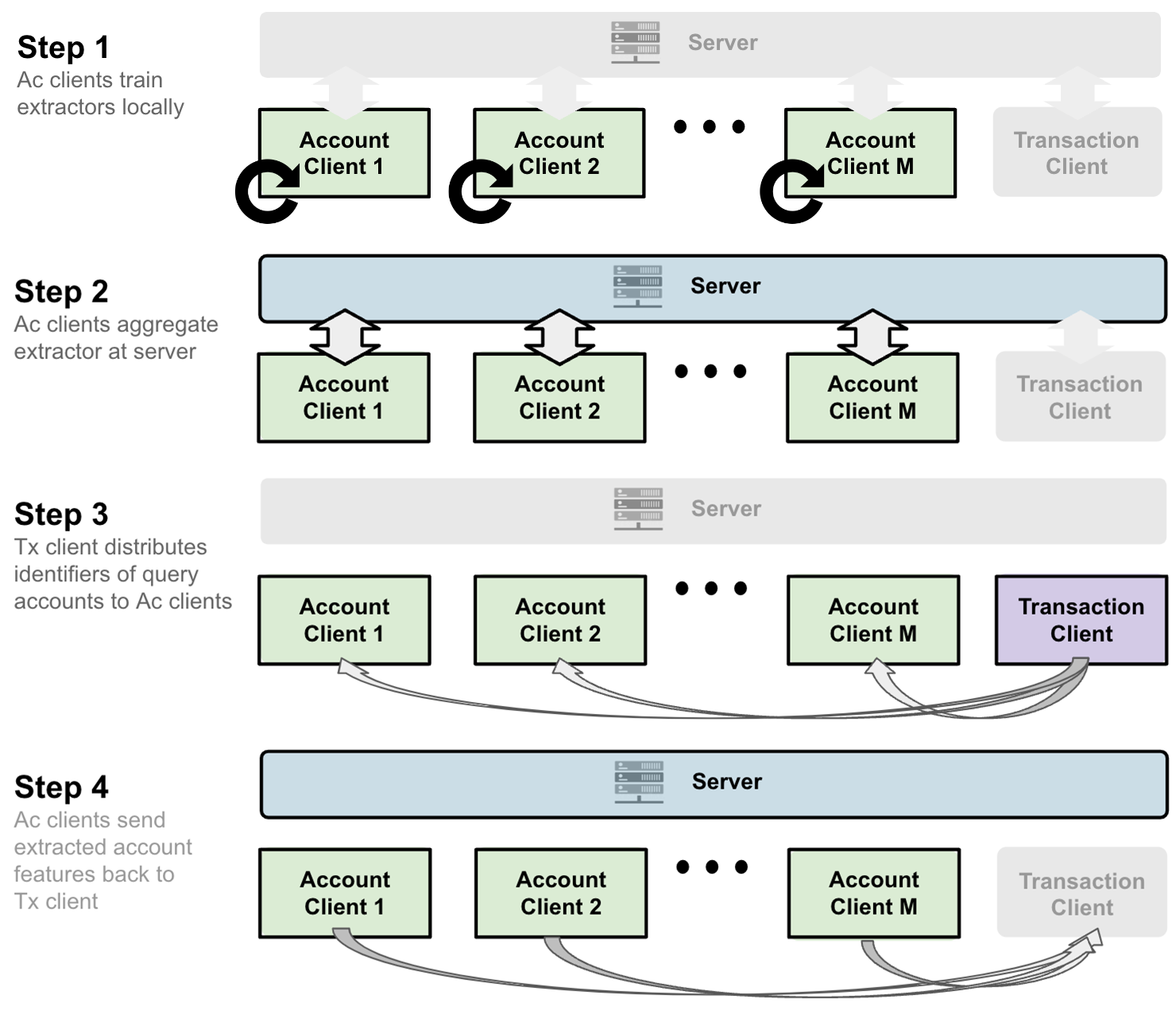} 
    \vspace{-0.05in}
    \caption{Communication protocol and information flow between the clients and the server in our proposed HyFL. (1) The account clients train the feature extractors locally. (2) The account clients send the extractors to the models and receive the aggregated model from the server. (3) The transaction client requests account meta information from the account clients by sending account identifiers. (4) The account clients send the features back to the transaction client.}
    \label{fig:framework}
    \vspace{-0.1in}
\end{figure*}

\noindent\textbf{Three Phases in the Training Stage. }
The proposed learning framework has three phases in the training stage:
\begin{enumerate}[leftmargin=*]
    \item (\textbf{Feature learning}) First, each account client trains a model to extract the feature embedding of the stored accounts data.
    Since there is no label stored in an account client, an auto-encoder is trained in a self-supervised fashion.
    Define the encoder and decoder as $h(\cdot)$ and $h'(\cdot)$.
    With a global initialization of $h_\phi, h'_{\psi}$, each client will train the models locally by
    \begin{align}
        \phi_m, \phi'_m = \arg \min\nolimits_{\phi, \phi'} \sum\nolimits_{i=1}^{N_m} \left\| h'_{\phi'}(h_\phi(\mathbf{x}_i)) - \mathbf{x}_i \right\|_2^2.
        \label{eq:train encoders}
    \end{align}
    When the number of account clients is very large, federated learning of the model has a large communication overhead.
    To reduce the overhead, the parameters of auto-encoders are aggregated in the server only once after the training in each account client: 
    \begin{align}
        \phi = {1\over M} \sum\nolimits_{m=1}^M \phi_m,~ \phi' = {1\over M} \sum\nolimits_{m=1}^M \phi'_m.
        \label{eq:aggregate encoders}
    \end{align}
    Then the aggregated auto-encoder will be broadcast back to all the account clients.

    \item (\textbf{Extract features}) Second, the account clients use the trained auto-encoder to extract the feature embedding of their local account data as $\mathcal{F}_m = \{h_\phi(\mathbf{x}_i)\}_{i=1}^{N_m}$. 
    Then, the transaction client will request the corresponding feature embeddings by querying account numbers.

    \item (\textbf{Classifier training}) 
    On receiving the embeddings, the transaction client will ensemble the account features with the transaction features as $\{([\mathbf{x}_{i,m_1}, \mathbf{x}_{i,m_2}, \mathbf{x}_{i,M}], y_i)\}$, where $\mathbf{x}_{i,m_1}$, $\mathbf{x}_{i,m_2}$ are the two account embeddings associated with the transactions.
    To protect data privacy from model inversion and membership inference, Gaussian noise is added to the concatenated features. 
    A high-capacity model such as XGBoost~\cite{chen2016xgboost} or neural networks (denoted as $f(\cdot)$) is then trained for predicting $y_i$, i.e., $f([\mathbf{x}_{i,m_1}, \mathbf{x}_{i,m_2}, \mathbf{x}_{i,M}])$. 
\end{enumerate}

Although our framework follows the client-server paradigm, where a dedicated server communicates with all clients, our framework can also work in a peer-to-peer (P2P) paradigm without a server: feature embeddings can be directly sent from the account clients to the transaction client.

%% file: sec/4_ThreatModel.tex
\subsection{Threat Model and Privacy Risk Sources}

The successful design of privacy-awareness machine learning systems depends on a realistic threat model and analysis of possible risk sources.
Based on our proposed HyFL framework, we elaborate on the threat models.
We first introduce four types of possible attacks. 
In the threat model, the privacy risk arises from the communications between any two nodes which could be of either the same or different types. 
We also discuss the privacy risks at the inference stage in addition to the communication during training.

\noindent\textbf{Privacy Risk Sources.}
The privacy risks of the proposed framework are four-fold. 
First, \emph{gradient inversion} targets recovering input data from a trained model. Sensitive information, such as data points, may be recovered from model parameters or gradients via this attack.
Another attack on the final trained model is the \emph{membership inference}, which infers a specific data point in the training set by evaluating the difference between the distribution of training data and test data. 
By stealing the latent correlations among features, \emph{attribute inference} can infer private information. 
There are also risks from unreliable units in the framework, where the \emph{feature leakage} can happen to extracted features. 
We give a formal definition to each of these four attacks as follows:
\begin{itemize}[leftmargin=*]
    \item \emph{Gradient inversion} is defined as the data reconstruction on the server-end with the support of the model parameters and gradients from the clients \cite{zhu2019dlg, geiping2020inverting}. Formally, given a model parameterized by $\theta$, the attacker makes use of the ground-truth gradients $g^*$ of mini-batch from a client to recover the original data in the client by matching the gradient on parameterized data $x$ in terms of the cosine similarity:
    \begin{equation*}
        \max_{x} \frac{\nabla^T\mathcal{L}_{\theta}(x^*;\theta) \cdot \nabla \mathcal{L}_{\theta}(x; \theta)}{\|\nabla\mathcal{L}_{\theta}(x^*;\theta)\| \cdot \| \nabla\mathcal{L}_{\theta}(x;\theta) \|},
    \end{equation*}
    where $x^*$ is the ground-truth data batch and $\mathcal{L}$ is the loss function. 
    The cosine similarity between the ground truth and estimated gradients is the objective of model inversion. 
    Since this objective function is differentiable, the Adam optimizer, which is a strong and commonly-used optimizer in deep learning, is typically utilized to maximize the recovered data until convergence.
    In federated learning, the gradients are typically replaced by the model update. 
    \item \emph{Membership inference} is first proposed in \cite{shokri2017membership}, which uses a binary classifier to predict whether a given data point is in the training set of the target model. 
    Specifically, the attacker has its own dataset, which may not contain labels. The attacker will first collect the predicted labels or logits of part of our own dataset using the target model from the API and then train our own model with the labeled dataset.
    Note that in FL, model parameters are usually released to the public and are considered to be a white-box setting. 
    Then we have a model to mimic the target model, a labeled dataset to train the model, and an unlabeled dataset not adopted for training. 
    The attacker can then use these two datasets to train a binary classifier that predicts the attendance of the data point in the original training set.
    \item \emph{Attribute inference} is the attack where the attacker uses a model to infer missing information of a data point from its incomplete information \cite{yeom2018privacy}. For instance, with tabular financial data, the adversary may have only part of the elements of each item. The attribute inference is to optimize the missing part towards the minimum loss.
    \item \emph{Feature leakage} is another potential privacy risk in FL that some unreliable modes may leak the features of data to other third parties. Despite the fact that feature leakage does not lead to direct data leakage, malicious third parties can use such features to adopt and support other kinds of privacy attacks like membership inference and attribute inference.
\end{itemize}

\noindent\textbf{Threat Model.}
A majority of FL systems focus on the \emph{honest but curious} nodes, i.e., one type of node will try to recover the data from the other two types of nodes, but it will not modify the model or features~\cite{semihonest}. 
In this work, we consider a comprehensive threat model by further assuming that the nodes of different types are \emph{potentially} honest but curious. 
Benefit from our framework, there is no direct communication between the transaction client and the server, so we can only investigate potential privacy risk sources from communications between two combinations of nodes:

\begin{enumerate}[leftmargin=*]
    \item During the communication between the aggregation server and account clients, model updates from account clients are sent to the server, which leads to direct gradient inversion risk as well as the membership inference.
    Also, optimization-based methods can be utilized to estimate private attributes with the support of some possible public data, which is the risk of attribute inference.
    \item During the communication between the account clients and the transaction client, features from the account clients will be sent to the server, which raises the problem of feature leakage. 
    Since the transaction has no access to the account clients' model, there is no risk of either model inference or attribute inference. 
    However, with the features from the account clients, membership can be inferred by the transaction client.
    \item The final risk lies in the transaction client. In the inference stage, attackers can utilize the predictive model in the transaction client to complete membership inference or attribute inference.
\end{enumerate}

\noindent\textbf{Privacy Risks in Both Stages.} 
In horizontal federated learning, each client can make predictions locally once trained well in the training stage~\cite{t2020personalized, fallah2020personalized} so the privacy risk decreases during the inference stage.
In our setting, however, the account meta-information, such as historical account activities and flags stored in the account clients, needs to be accessed in both training and inference stages, which means the transaction client cannot access such information directly. 
Moreover, this indicates that even in the inference stage, all the nodes are involved in the prediction, which is different from the horizontal framework. 
Even worse, we consider that attackers can steal data in the inference stage by utilizing the model inference or membership inference attack on the trained classifier. 
As such, the challenge arises that we need to protect privacy in both stages. We show that the privacy risks of each communication discussed above are similar for both stages so that we can ensure privacy in both stages by targeting on the three privacy risk sources.

\vspace{+0.05in}
\noindent\textbf{Solutions.} 
Targeting on the three privacy risk sources, we propose to adopt various defense mechanisms in our framework to protect privacy. 
    (1) To defend the model inversion and membership inference attack, we add Gaussian noise to the data in the transaction clients before training the classifier, which confuses the inversion or inference attackers from the real features. 
    (2) To defend the attribute inference attack, we keep the encoder model only to account clients and away from the transaction client such that the transaction client cannot recover the sensitive attribute.
    (3) We also leverage encryption during the communication so that the server cannot recover the original features from model parameters.

%% file: sec/5_System.tex
\section{System Implementation}
\label{sec:system}

In this section, we describe the technical details of our novel Hybrid Federated Learning (HyFL) framework. We introduce the training and inference stages, approaches to preserve privacy during the two stages, and techniques to increase the scalability of our framework and make it feasible. 
We first introduce a vanilla framework as the backbone of our framework. Then we enhance the vanilla framework with privacy-preserving approaches.%

\vspace{+0.1in}
\noindent\textbf{Hybrid Federated Learning Backbone.}
At the beginning of our framework, the account clients train their feature extractors locally. 
Then the server collects and aggregates these trained extractors and sends the aggregated extractor back to the account clients.
Since a portion of the features and the labels are stored in the transaction client, both the training and inference stages start from the transaction client. 
We can either directly use the original features or use a feature extractor to extract features, depending on the size of the features used. If the feature size is small, then we can work with the original feature space. Otherwise, if high-dimensional data such as text and images are used, then we recommend using a feature extractor for dimension reduction.   
Next, the original or extracted features will be sent to the server. The transaction client will use account identifiers to query for the corresponding features stored in the account clients. 
When an account client receives a query from the transaction client for a specific account, it searches for the corresponding account features (e.g., flags) and uses a feature extraction model to generate the intermediate features.
Then the extracted features will be sent back to the transaction client to predict the final label. 
Due to the lack of labels in account clients, we propose to use an unsupervised model as our feature extractor, and the autoencoder~\cite{bank2020autoencoders, wang2014generalized} is an ideal instantiation. 
Our final implementation uses XGBoost~\cite{chen2016xgboost} as the classifier in the transaction client for its strong performance in classification tasks.
This concludes the backbone of our framework, which we call the vanilla HyFL framework.
The framework of training and inference stages is summarized in Algorithm~\ref{algo:naive}.
Note that in the vanilla HyFL framework, the server only has the model of feature extractors, but no features from the account clients.
On the other hand, the transaction client only has the extracted features but not the encoders, so it cannot recover the data from the features only.

\begin{algorithm}[t]
\renewcommand{\algorithmicrequire}{\textbf{Input:}}
\renewcommand{\algorithmicensure}{\textbf{Output:}}
\caption{Vanilla HyFL Framework without Privacy Protection.}
\begin{algorithmic}
\label{algo:naive}
\REQUIRE Transaction features without account meta information in the transaction (Tx) client%
\REQUIRE Accounts associated with account meta information in account (Ac) clients%
\STATE \textbf{Ac clients} locally train $\phi_m, \phi'_m$ with Eq.~(\ref{eq:train encoders})
\STATE \textbf{Server} aggregates $\phi_m,\phi'_m$ with Eq.~(\ref{eq:aggregate encoders}) and returns the extractors to Ac clients
\WHILE{Training or Inference}

\STATE \textbf{Tx client} sends accounts to account clients
\STATE \textbf{Ac clients} (1) query and extract features according to the accounts, (2) send extracted features to transaction client
\STATE \textbf{Tx client} receives the extracted features and makes a prediction.

\ENDWHILE
\end{algorithmic}
\end{algorithm}

\vspace{+0.1in}
\noindent\textbf{Privacy Preserving Enhancement.}
To tackle the three privacy risk sources, we enhance our framework through the lens of several privacy-preserving methods, including differential privacy mechanisms, feature sharing, and homomorphic encryption. 
The risk sources are first analyzed for each communication, then targeted solutions are identified correspondingly.
\begin{enumerate}[leftmargin=*]
    \item The first risk source is the model sharing from account clients to the server, which induces the risk of inference attacks and inversion attacks. 
    To address this issue, we adopt encryption to encrypt the model parameters before sending them to the server. Without the private key, the server cannot decrypt the data but can still aggregate the encrypted parameters.
    \item Another risk source is the feature sharing from the account clients to the transaction client.
    To alleviate the risk of attribute inference attacks and feature leakage, instead of directly sending the original account meta information themselves, a feature extractor is adopted by the account clients.
    Without access to the model in the account clients, the transaction client cannot recover the original account data with only the extracted features.
    \item The final risk source is that in the inference stage, the predictive model in the transaction client may be used to infer the membership or private attribute. 
    To mitigate this problem, we add privacy-protecting Gaussian noise to the data in the transaction client before training the classifier. 
    Note that our data is in tabular form. In that case, unlike image data, some little perturbation can lead to fidelity.
    Normalization to the standard normal distribution is utilized as well before the noise injection, which changes the distribution of original data, making it even more difficult to recover the data for the attackers.

\end{enumerate}

Strengthening our vanilla HyFL framework with the above privacy-preserving enhancement and feasible scalability improvement, we illustrate the final HyFL framework in Algorithm~\ref{algo:HFL}.
The key difference versus Algorithm~\ref{algo:naive} is the encryption and noise mechanism for privacy protection.

\begin{algorithm}[t]
\renewcommand{\algorithmicrequire}{\textbf{Input:}}
\renewcommand{\algorithmicensure}{\textbf{Output:}}
\caption{The proposed HyFL framework.}
\begin{algorithmic}
\label{algo:HFL}
\REQUIRE Transaction features without account meta information in the transaction (Tx) client%
\REQUIRE Account meta information in account (Ac) clients%

\STATE \textbf{Ac clients} (1) locally train $\phi_m,\phi'_m$ with Eq.~\ref{eq:train encoders} and send to the server%
\STATE \textbf{Server} aggregates $\phi_m,\phi'_m$ with Eq.~\ref{eq:aggregate encoders} and returns to the Ac clients

\WHILE{Training}
\STATE \textbf{Tx client} sends account identifiers to the Ac clients
\STATE \textbf{Ac clients} query and \textit{encrypt} features according to the accounts and send them to the Tx client
\STATE \textbf{Tx client} (1) receives and \textit{decrypts} the features, (2) adds Gaussian noise into the aggregated features, (3) trains a classifier $f$ \textit{with aggregated features}

\ENDWHILE

\WHILE{Inference}
\STATE \textbf{Tx client} sends account identifiers to Ac clients
\STATE \textbf{Ac clients} query and \textit{encrypt} features according to the accounts identifiers and send to the Tx client
\STATE \textbf{Tx client} (1) receives and \textit{decrypts} the features; (2) predicts the label $\hat{y}$ with $f$.
\STATE \textbf{Tx client} Outputs the prediction label $\hat{y}$
\ENDWHILE
\end{algorithmic}
\end{algorithm}

\begin{figure}[t]
	\centering
	\includegraphics[width=0.5\textwidth]{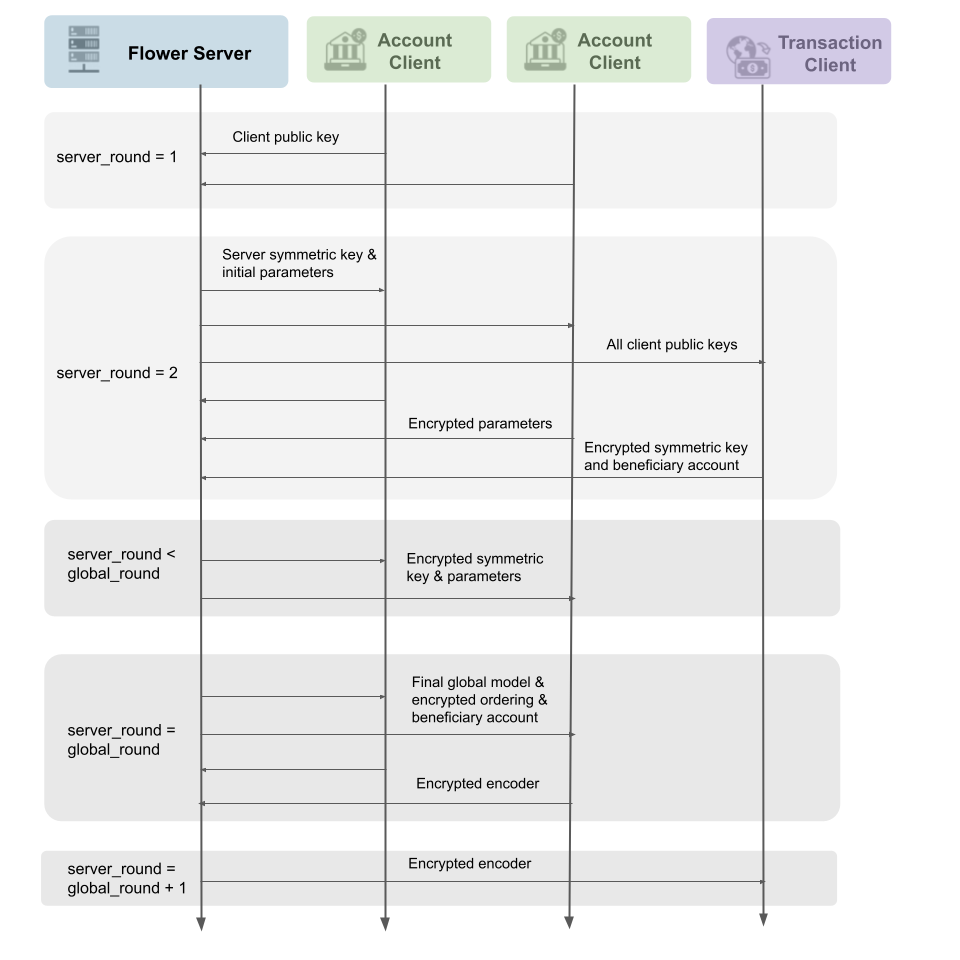}
 \vspace{-0.35in}
	\caption{Overview of the communication flow of the proposed HyFL framework in the training stage.}
	\label{fig:train}
\end{figure}

\begin{figure}[t]
	\centering
	\includegraphics[width=0.5\textwidth]{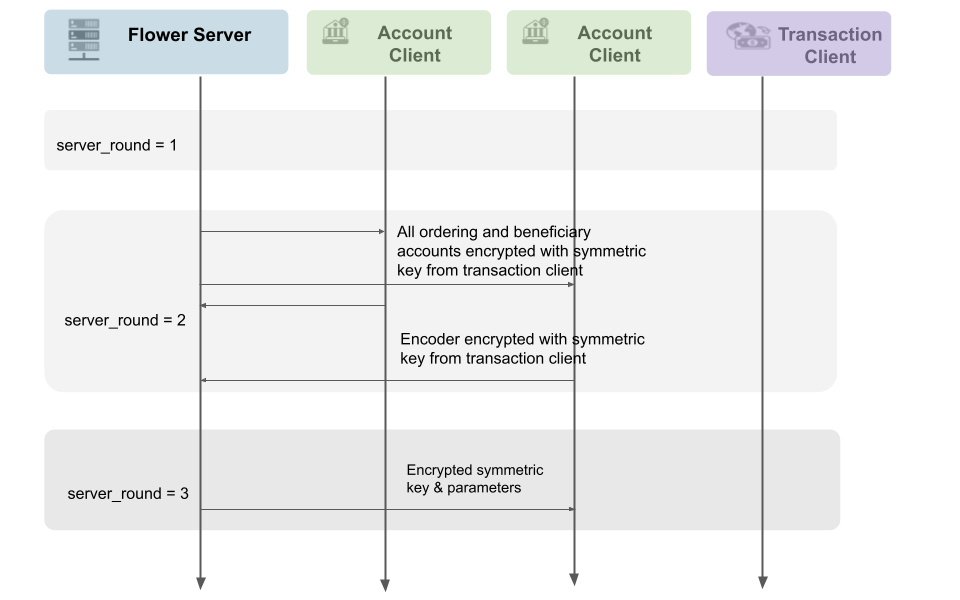}
 \vspace{-0.25in}
	\caption{Overview of the communication flow of the proposed HyFL framework in the inference stage.}
	\label{fig:test}
\end{figure}

\noindent\textbf{Summary of Communication Flows.} Considering the aforementioned improvements and privacy protection, we built an implementation of the proposed HyFL based on the Flower~\cite{beutel2022flower} project. 
We summarize the complete communication flows of  our implementation in the training and inference stages in Fig.~\ref{fig:train} and Fig.~\ref{fig:test}, respectively. An encryption key exchange is involved during the initialization of the training stage, which will also be used in the testing stage.

%% file: sec/6_ExperimentalResults.tex
\section{Experimental Results}
\label{sec:experiments}

\subsection{Experiment Setup}
\subsubsection{Dataset}
We use the synthetic dataset provided by SWIFT.
The data we use is in the form of tables with a limited number of features but of large size, which is suitable for the model training and queries across different nodes in our framework. 
Since we have billions of samples but the number of features is small, it yields a high risk of overfitting if we use a neural network as the classifier, which is backed up by our conducted experiments. Yet this kind of dataset provides a unique opportunity for the model such as XGBoost. 

Specifically, the size of the training set is approximately three million, with the proportion of positive and negative samples being $1:100$. The size of the test set is one-quarter of the training set. 
The data stored in the transaction client consists of seven features, such as the transaction amount, transaction frequency, and the average amount for the currency used in each transaction. 
While the account meta information is the status of the receiver and sender in each transaction, which is simple yet important to determine whether a transaction is a crime or not. 
Such status ranges from $0$ to $11$, which we use an encoder to transform into continuous features. 
For the number of account clients, we consider a wide range from $1$ to $200$, where we analyze the impact of the number in Sec~\ref{sec:impact of account client}. 

Note that besides structured data, the proposed HyFL framework can also handle unstructured data like images and text, with various model architectures such as convolution neural networks, recurrent neural networks, and even Transformers. 
Such design allows us to consider more versatile crime detection scenarios, for example, imaging or even video data types in the Know Your Customer (KYC) information can be considered by financial crime detection. 
To apply our framework in these scenarios, one should first choose the model architecture of feature extractors in the account clients. 
Due to the limited number of features and strong performance of XGBoost, there is no need for us to use a feature extractor in the transaction client. However, the user can explore different kinds of models as the feature extractor based on the data type. 
For example, if the data is images, the user may choose convolution neural networks as the extractor. Then the account clients can extract features and send them to the server to help with the training of the classifier.
Moreover, we demonstrate that our framework is a general paradigm for hybrid federated learning, which can be utilized even when the data form in two types of clients is different. For instance, image data in the account clients can be used to enhance the training of text data in the transaction client, as long as an appropriate feature extractor is utilized for each data type.

\subsubsection{Evaluation Metrics and Implementation}
Due to the data imbalance, we use AUCPR as the main metric to evaluate model performance, which is defined as $\sum_k(R_k-R_{k-1})P_k$, where $R_k$ and $P_k$ are the recall and precision under the $k$-th threshold. 
We also report precision, recall and F1-score in our results, to make a comprehensive analysis of our proposed framework.
To evaluate the trade-off between privacy and utility, we use the variance of the noise as the privacy budget, as well as the average $l_2$ norm and average cosine similarity between original data and noise-added data.

\subsection{Comparison with Baselines}
In this section, we compare our framework with the baselines, to evaluate its effectiveness.
We set $0.01$ as a reasonable noise variance according to the Fig.~\ref{fig:impact of variance}. 
Since we have both vertical and horizontal settings in our framework, current frameworks cannot be utilized to evaluate our proposal. 
Thus, we consider the following two settings as our baselines:
(1) Centralized setting where all the account data and transaction data are stored in a single node, known as the centralized scenario. 
(2) Vanilla HyFL as in Alg~\ref{algo:naive} where the setting is the same as our proposed framework but without noise. 

To conduct a comprehensive evaluation, we compare our framework with four typical models for binary classification tasks: XGBoost present for tree models, Support Vector Machine (SVM)~\cite{boser1992training} present for linear models, Logistic Regression (LR)~\cite{lavalley2008logistic} present for simple non-linear models, and multi-layer perceptron (MLP)~\cite{gardner1998artificial} present for deep non-linear models.

The results for three settings are shown in Table~\ref{tab:baseline}. 
The performance of SVM is the worst across the three settings, which implies that the linear classification model is not suitable for such a task with tabular data, because such data is not linearly separable. 
Similarly, although LR has some non-linear properties from the sigmoid function, it still cannot achieve satisfactory performance due to the linear combination of features before the sigmoid function.
On the other hand, MLP shows great improvement with the support of strong power to extract useful features from raw data. 
Surprisingly, XGBoost achieves the best performance across all three settings. 
One possible reason is that as the number of estimators in XGBoost increases, the complex relationship between features can be extracted. 
In that case, XGBoost can also construct a strong non-linear map from features to the predicted label just like MLP. 

From the perspective of different frameworks, we can observe that the vanilla setting has a similar performance to the centralized setting as shown in Table~\ref{tab:baseline}. 
Note that the vanilla setting has the same framework backbone as HyFL setting, except for the noise injection and encryption. 
Based on this perspective, the similar performance between vanilla and the centralized setting implies that our framework backbone is effective as the centralized setting.
Table~\ref{tab:baseline} shows that the noise indeed decreases the performance, but it still presents a reliable utility with a relatively high AUCPR.
We further analyze the impact of noise variance in Sec~\ref{sec:impact of noise}.

\begin{table}
\begin{tabular}{l|llll}
\hline & XGBoost & SVM & LR & MLP \\ \hline 
& \multicolumn{4}{c}{\cellcolor{gray!15} Centralized setting} \\ \hline
Precision & 0.97 & 0.50 & 0.97 & 0.99 \\ %
Recall & 0.79 & 0.46 & 0.61 & 0.67 \\ %
F1 & 0.86 & 0.39 & 0.68 & 0.76 \\ %
AUCPR & 0.7037 & 0.0011 & 0.2976 & 0.5608 \\ \hline
& \multicolumn{4}{c}{\cellcolor{gray!15}Vanilla HyFL} \\ \hline
Precision & 0.98 & 0.50 & 0.97 & 0.98 \\ %
Recall & 0.79 & 0.37 & 0.61 & 0.71 \\ %
F1 & 0.86 & 0.13 & 0.68 & 0.79 \\ %
AUCPR & 0.7075 & 0.0009 & 0.2977 & 0.5392 \\ \hline
& \multicolumn{4}{c}{\cellcolor{gray!15}HyFL} \\ \hline
Precision & 0.98 & 0.50 & 0.97 & 0.95 \\ %
Recall & 0.76 & 0.37 & 0.61 & 0.70 \\ %
F1 & 0.83 & 0.13 & 0.68 & 0.78 \\ %
AUCPR & 0.6839 & 0.0009 & 0.2975 & 0.5438 \\ \hline
\end{tabular}
\caption{Performance of different classifiers under three different settings.}
\label{tab:baseline}
\vspace{-10pt}
\end{table}

\subsection{Ablation Study}
\subsubsection{Impact of Sampling}
In this section, we analyze the impact of different sampling methods used to solve data imbalance.
The proportion of the negative and positive data in our training set is about $1:783.73$. 
The typical method to solve such data imbalance is to resample the training set so that the number of positive and negative samples is balanced. 
Another typical method is to reweight the losses of positive and negative samples.
We evaluate four methods to evaluate the impact of imbalance-target methods:
(1) RandomUnder, which randomly under-sample the negative samples.
(2) RandomOver, which randomly over-sample the positive samples
(3) SMOTE, which uses KNN to generate synthetic data to augment positive samples.
(4) Reweight, which assigns a higher weight to the loss of the positive samples.
Surprisingly, we find the model performance is hampered by all the resampling methods. 
Due to the great data imbalance, such resampling methods will drop a large number of negative samples, which leads to a catastrophe in model performance. 
Instead, reweighting only suffers from a small loss of model performance since it does not drop any negative samples but forces the model to focus more on the positive samples.

\begin{table}[]
\resizebox{0.47\textwidth}{!}{%
\begin{tabular}{l|llll}
\hline
& RandomUnder & RandomOver & SMOTE & Reweight \\ \hline
Precision & 0.55 & 0.51 & 0.51 & 0.51 \\ %
Recall & 0.80 & 0.83 & 0.81 & 0.88 \\ %
F1 & 0.58 & 0.49 & 0.51 & 0.52 \\ %
AUCPR & 0.4712 & 0.4717 & 0.4572 & 0.67 \\ \hline
\end{tabular}
}
\caption{Evaluation of imbalance-target methods.}
\label{tab:impact of sampling}
\vspace{-20pt}
\end{table}

\begin{figure}[htbp]
\vspace{-20pt}
\centering
    \subfloat[AUCPR vs. Variance]{%
        \includegraphics*[width=0.35\textwidth, height=0.45\linewidth]{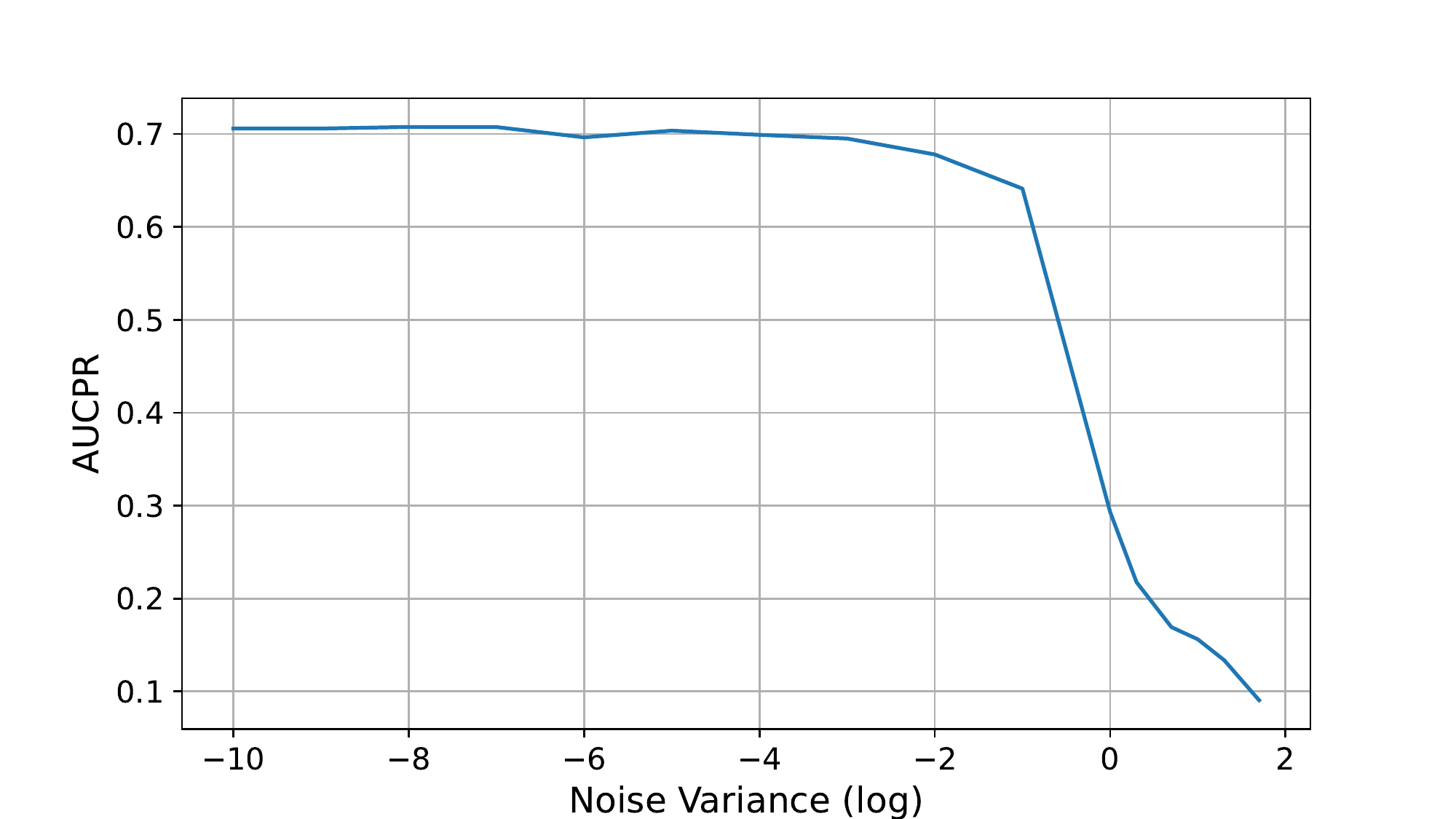}
        \label{fig:AUCPR vs. Variance}
    }     
    \begin{center}
    \end{center}
    \vspace{-15pt}
    \subfloat[Norm / CosSim vs. Variance]{%
        \includegraphics*[width=0.35\textwidth, height=0.5\linewidth]{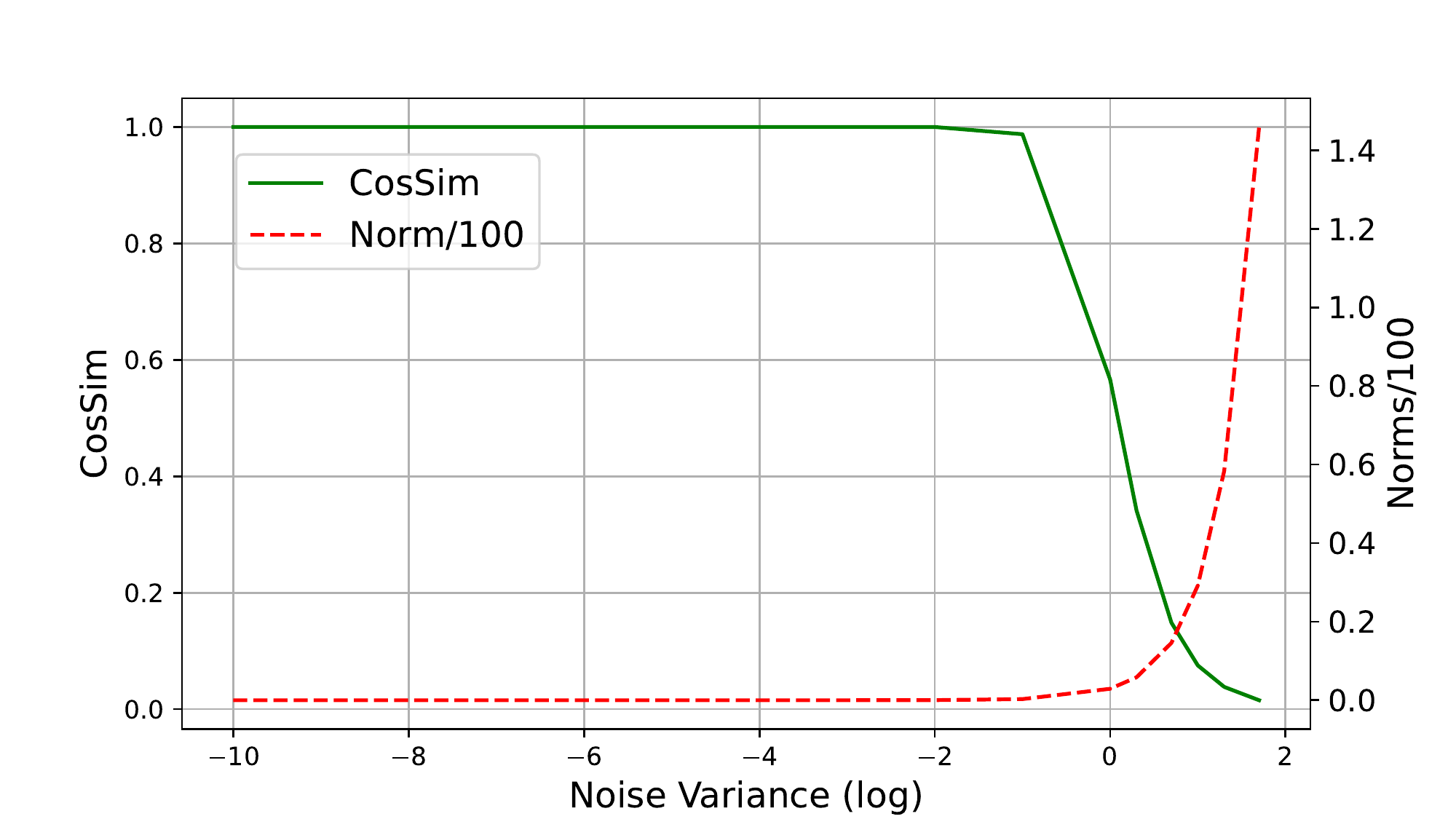}   
        \label{fig:CosNorm vs. Variance}
    }
    \vspace{-0.15in}
    \caption{The impact of noise variance added into the training data on three factors: AUCPR, Norm, and CosSim. AUCPR is to evaluate model performance on classification. The Norm is calculated as the average norm of noise added to each data point. The CosSim is the average cosine similarity of noisy samples and original samples.}
    \label{fig:impact of variance}
\end{figure}

\subsubsection{Impact of Communication Interval}
In this section, we analyze the impact of the number of communication rounds between the bank clients and the server, as well as that of the communication interval.
We set the total number of epochs as $50$, where is the multiplication of the communication interval ($I$) and the communication round number ($R$). 
Specifically, we consider four settings with different combinations of $I$ and $R$.
We find if the number of the account clients is small, $I$ and $R$ only have little influence on the performance of the final model. 
Hence, in this section, we use $100$ account clients to evaluate the impact of $I$ and $R$, presented in Table~\ref{tab:impact of round}.
It is shown that more frequent communication leads to better model performance. 
However, due to the overhead of frequent communication, a trade-off is necessary between the communication cost and the model performance.

\begin{table}[]
\begin{tabular}{l|llll}
\hline
& I1-R50 & I5-R10 & I10-R5 & I50-R1 \\ \hline
Precision & 0.97 & 0.97 & 0.97 & 0.97 \\ %
Recall & 0.79 & 0.79 & 0.71 & 0.71 \\ %
F1 & 0.86 & 0.86 & 0.79 & 0.79 \\ %
AUCPR & 0.7037 & 0.7037 & 0.5344 & 0.5344 \\ \hline
\end{tabular}
\caption{Performance evaluation of the number of communication rounds.}
\label{tab:impact of round}
\vspace{-15pt}
\end{table}

\begin{table}[]
\begin{tabular}{l|lllll}
\hline
& 1 & 10 & 50 & 100 & 200 \\ \hline
Precision & 0.97 & 0.97 & 0.97 & 0.97 & 0.97 \\ %
Recall & 0.79 & 0.79 & 0.71 & 0.71 & 0.71 \\ %
F1 & 0.86 & 0.86 & 0.79 & 0.79 & 0.79  \\ %
AUCPR & 0.7037 & 0.7037 & 0.5344 & 0.5344 & 0.5344 \\ \hline
\end{tabular}
\caption{Impact of account client number on the performance.}
\label{tab:impact of bank num}
\vspace{-18pt}
\end{table}

\begin{table}[]
\begin{tabular}{l|llll}
\hline
& 0.5 & 0.1 & 0.01 & 0.002 \\ \hline
Precision & 0.97 & 0.94 & 0.97 & 0.59 \\ %
Recall & 0.79 & 0.75 & 0.71 & 0.51 \\ %
F1 & 0.86 & 0.82 & 0.79 & 0.51 \\ %
AUCPR & 0.7037 & 0.6068 & 0.5344 & 0.0534 \\ \hline
\end{tabular}
\caption{Impact on the model performance of the data size.}
\label{tab:impact of data size}
\vspace{-20pt}
\end{table}

\subsubsection{Impact of Noise}
\label{sec:impact of noise}
We give an analysis of the model performance w.r.t. different noise variances, shown in Fig.~\ref{fig:impact of variance}.
We utilize Gaussian noise with the mean as zero and analyze the model performance with different variances.
To study the impact of noise variance, we consider three factors: AUCPR, average norm, and average cosine similarity between original data and noise-injected data.
From Fig~\ref{fig:AUCPR vs. Variance}, we can see that noise with a small variance until $10^{-3}$ can enhance the model generalization, and hence the AUCPR can be improved. 
The norm and cosine similarity are close to zero and one, respectively, which indicates that there is little influence on original data with such small noise.
However, as the variance increases beyond $10^{-2}$, both the cosine similarity and AUCPR decrease significantly, and the norm increases greatly. 
Thus, the results present a trade-off between the privacy budget and the model utility.
It is shown that with the variance as $10^{-3}$, the AUCPR on test data peaks, which means a good generalization performance in the inference stage.
In this case, we can have a better trade-off between privacy and performance.

\subsection{Sensitivity Study}
\subsubsection{Impact of Account Client Number}
\label{sec:impact of account client}
In this section, we present the impact of the number of account clients on the model performance. 
We only change the account client number and maintain the other settings in this section. 
Based on the account client number, the account data is split randomly and used to train an auto-encoder locally in each account client.
Then the server aggregates all the auto-encoders from the account clients as the global auto-encoder.
The results are summarized in Table~\ref{tab:impact of bank num}.
As the number of account clients increases, the model performance tends to be stable, which implies that our framework is robust even with a large number of account clients. 
Yet the model can achieve a better performance when the account client number is small, we can still obtain utility as the account client number is large.

\subsubsection{Impact of Data Size}
In this section, we study the impact of the size of the training set. 
We randomly sample the training data from the original training set, based on the sampling ratio, where we maintain the relative proportion between positive and negative samples.
From Table~\ref{tab:impact of data size} we can see the model performance is destroyed by the decreasing data size. 
Note the number of positive samples is quite limited in our case. 
Hence, as we decrease the data size, the number of positive samples is getting smaller, which makes it even more difficult to learn from the positive samples.

%% file: sec/9_Conclusion.tex
\vspace{-0.1in}
\section{Conclusion}
In this paper, we tackled the challenges of the collaboration of multiple financial institutions for financial crime detection. 
We showed that existing FL paradigms could not be directly applied to solve this detection problem because of the way private data and features are distributed.
To address the challenges, we proposed a hybrid FL framework that allows sharing features and training the model in a privacy-preserving way. 
Empirical results showed the robustness and effectiveness of the framework under different scenarios. 
We believe the proposed solution can greatly transform the landscape of the financial system by better safeguarding them with a powerful financial crime detection tool.